\documentclass[letterpaper, 10 pt, conference]{ieeeconf}  % Comment this line out if you need⁸

\IEEEoverridecommandlockouts                              % This command is only needed if 
\usepackage{amsmath} % assumes amsmath package installed

\usepackage{cite}
\usepackage{xcolor}
\usepackage{graphicx}
\usepackage{subcaption}
\usepackage{booktabs}
\usepackage[table]{xcolor}

\title{\LARGE \bf
Robust Graph Matching through Semantic Relationship Generation for SLAM
}

\author{David Perez-Saura$^{1,2}$, Jose Andres Millan-Romera$^{1}$, Miguel Fernandez-Cortizas$^{1}$, \\
 Holger Voos$^{1}$, Pascual Campoy$^{2}$, and Jose Luis Sanchez-Lopez$^{1}$% <-this % stops a space
    \thanks{This work was supported by the Fonds National de la Recherche of Luxembourg (FNR), under the projects C22/IS/17387634/DEUS and 17097684/RoboSAUR, and by the European Union's Horizon Europe projects SHEREC “Safe Healthy and Environmental Ship Recycling", ref: 101136056, and “CORESENSE: A Hybrid Cognitive Architecture for Deep Understanding”, ref: 101070254.
    It has also been supported by the project CPP2022-009933 funded by MICIU/AEI /10.13039/501100011033 and by European Union NextGenerationEU PRTR, 
    and the work of the first author is supported by the Program for Technical Assistants PTA2021-020671, funded by the Spanish Ministry of Science and Innovation.}
    \thanks{$^{1}$ Automation and Robotics Research Group, Interdisciplinary Centre for Security, Reliability and Trust, University of Luxembourg, 1855 Luxembourg, Luxembourg. Holger Voos is also with The Faculty of Science, Technology and Medicine, University of Luxembourg, 1855 Luxembourg, Luxembourg. \tt{\{jose.millan, miguel.fernandez, holger.voos, joseluis.sanchezlopez\}}@uni.lu}
    \thanks{$^{2}$Computer Vision and Aerial Robotics Group at Universidad Politécnica de Madrid, 28040, Spain (CVAR-UPM) at the Centre for Automation and Robotics (UPM-CSIC). 
    \tt{\{david.perez.saura, pascual.campoy\}@upm.es}}
}
\begin{document}

\maketitle
\thispagestyle{empty}
\pagestyle{empty}

%%%%%%%%%%%%%%%%%%%%%%%%%%%%%%%%%%%%%%%%%%%%%%%%%%%%%%%%%%%%%%%%%%%%%%%%%%%%%%%%
\begin{abstract}

Graph-based representations such as Scene Graphs enable localization in structured indoor environments by matching a locally observed graph, constructed from sensor data, to a prior map. This process is particularly challenging in environments with repetitive or symmetric layouts, where structural cues alone are often insufficient to resolve ambiguities.

% Graph-based representations such as Scene Graphs enable structured localization in indoor environments, but matching these graphs remains challenging in the presence of repetitive or symmetric layouts. In such scenarios, structural cues alone are often insufficient to uniquely determine correspondences.

We propose a semantic-enhanced graph matching approach that explicitly models relations between detected objects and structural elements, such as rooms and wall planes. Objects are detected from RGB-D data and integrated into the graph, and their relations to structural elements are exploited to filter candidate correspondences prior to geometric verification, significantly reducing ambiguity and search complexity.

The proposed method is integrated within the iS-Graphs framework and evaluated in synthetic and simulated environments. Results show that semantic relations significantly reduce the number of candidate matches, improve computational efficiency, and enable faster convergence, particularly in symmetric scenarios where purely geometric approaches fail.

\end{abstract}

%%%%%%%%%%%%%%%%%%%%%%%%%%%%%%%%%%%%%%%%%%%%%%%%%%%%%%%%%%%%%%%%%%%%%%%%%%%%%%%%

\section{Introduction}

Mobile robots are increasingly deployed for inspection and monitoring tasks in structured indoor environments such as construction sites. To enable such applications, robots must accurately localize themselves within the environment and match their observations with the original architectural plans. 
To represent both within a unified framework, graph-based Scene Graphs~\cite{armeni20193dscenegraph} have recently emerged as an effective way to encode the structure of indoor environments. In particular, hierarchical representations such as Situational Graphs (S-Graphs) \cite{bavle2023sgraphs+} model the environment using semantic entities, including structural elements such as walls and rooms, together with robot poses and observations. Building on this idea, \textit{iS-Graphs} \cite{shaheer2023graph} leverages the topological and relational information to match site plans with the robot's observations. This match not only allows the localization with respect to the architectural plan, but also its use in the SLAM optimization, improving trajectory estimation and map reconstruction.

\begin{figure}[t]
      \centering
      \includegraphics[width=1.0\linewidth]{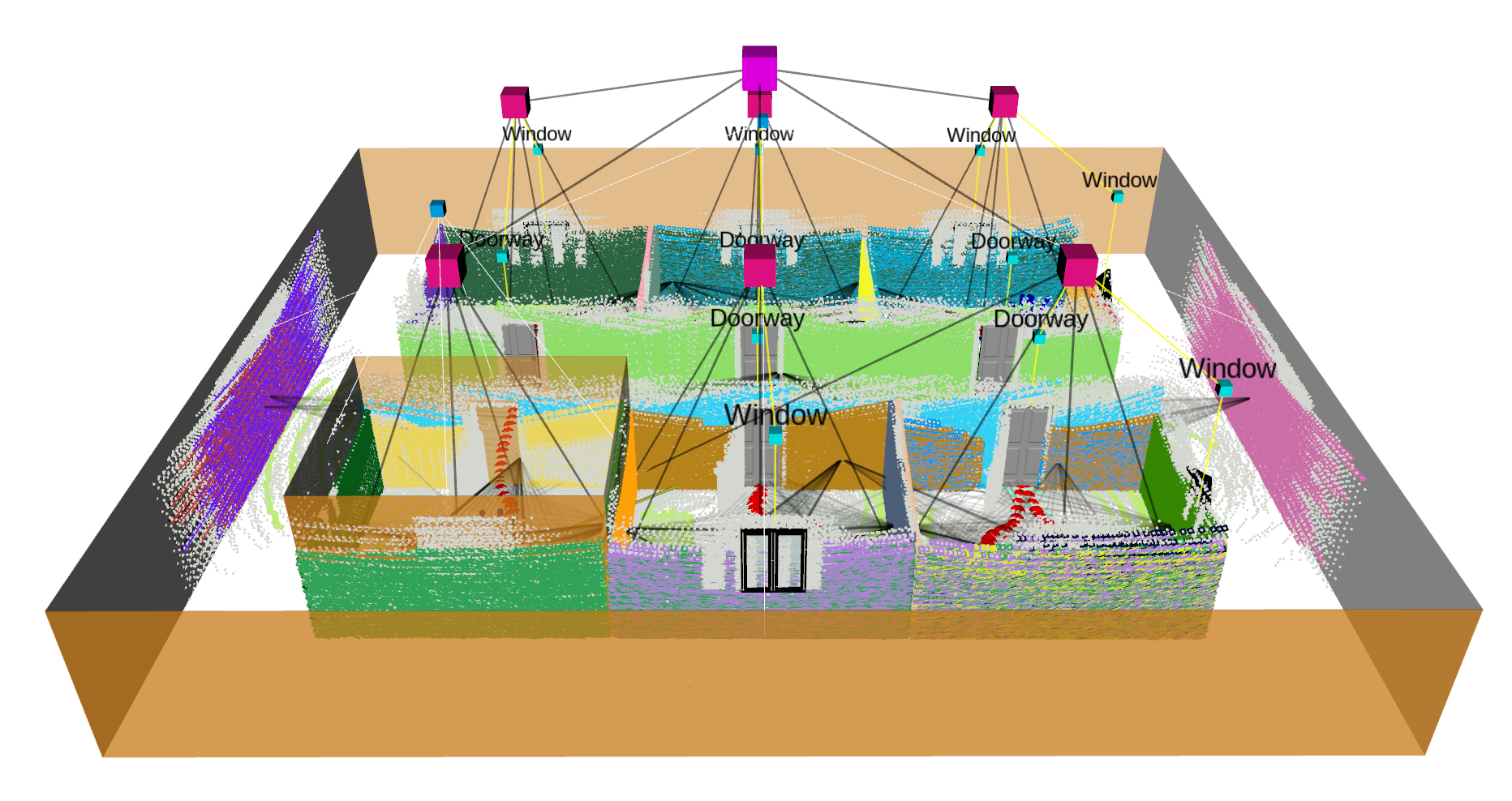}
      \caption{Semantic-enriched scene graph showing detected objects (e.g., windows, doorways) and their relations to structural elements within the environment. The generated object-semantic relations are integrated into both the S-Graph and A-Graph, enabling semantic-enhanced graph matching.}
      \label{figurelabel}
\end{figure}

However, relying only on the building structure to match the architectural graph and the observed graph
% Although hierarchy, graph topology, and geometric consistency checks can filter many incorrect matches, matching both graphs 
remains challenging in environments with repetitive or symmetric layouts. These approaches overlook other elements (such as doorways), which exhibit meaningful relations with structural elements and can help disambiguate correspondences that purely geometric constraints cannot resolve.

In this work, we address this limitation by detecting objects in the environment, incorporating them into both graphs, and exploiting their relations with structural elements to filter candidate correspondences prior to geometric verification. This enables early rejection of unfeasible candidates, which not only makes the correct match identifiable in environments with repeated structural layouts, but also significantly reduces the overall computation time.

\begin{figure*}[t]
    \centering
    \includegraphics[width=1.0\textwidth]{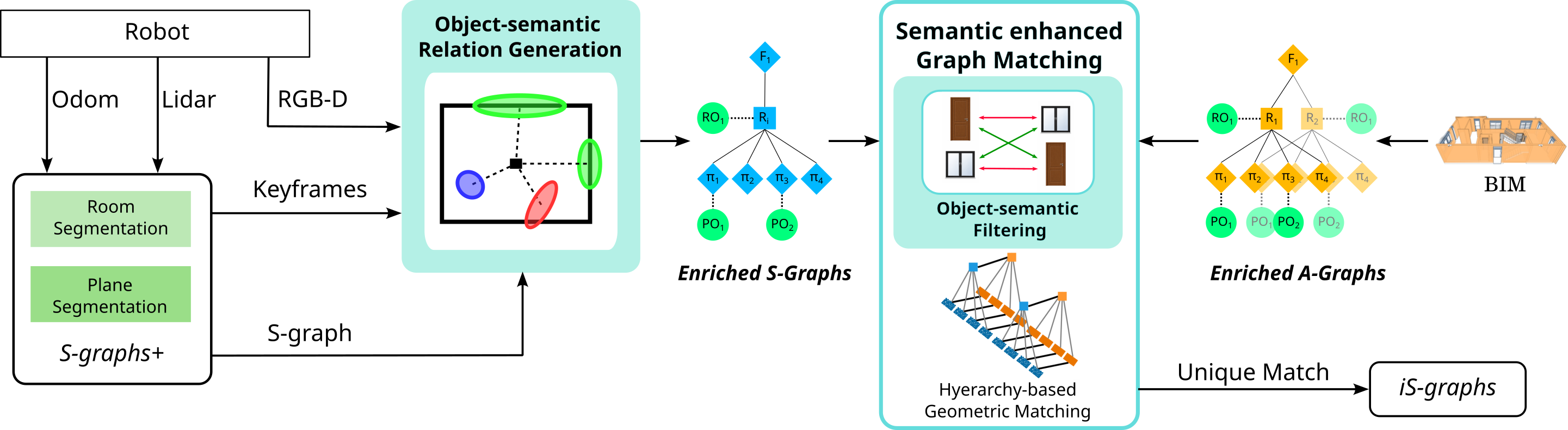}
    \caption{Overview of the proposed semantic-enhanced graph matching framework. An S-Graph is generated online, while a prior A-Graph is extracted from a BIM model. Detected objects from RGB-D data are incorporated to the S-graph through the object-semantic relation generator, which establishes relations between objects and structural elements. These relations enrich both graphs and are used in a semantic filtering stage to prune candidate correspondences before hierarchy-based geometric matching. The resulting unique match is then used for graph coupling within the \textit{iS-Graphs} framework for localization.}
    \label{fig:main_diagram}
    \vspace{-10pt}
\end{figure*}

The main contributions of this work are summarized as follows:

\begin{itemize}
\item A pipeline to detect and localize objects from multiple detections from RGB-D data.
\item A method to generate semantic relations between detected objects and structural elements such as rooms and wall planes within the S-Graph representation.
% \item A semantic-enhanced graph matching strategy that exploits these relations to filter candidate correspondences between A-Graphs and S-Graphs.
\item A semantic-enhanced graph matching strategy that exploits these relations to filter candidate correspondences between an observed graph and an architectural graph.
\item Integration of the proposed approach into the \textit{iS-Graphs} framework.
% \item Experimental validation demonstrating improved robustness of graph matching in structured indoor environments with repetitive layouts.
\end{itemize}

\section{RELATED WORK}

\subsection{3D Semantic Scene Graphs}

The use of semantic information has gained increasing attention to improve robustness and scalability in robotic localization. In this context, 3D scene graphs~\cite{armeni20193dscenegraph} provide a structured representation that integrates geometry, objects, and higher-level entities such as rooms.
Subsequent works extended this idea to hierarchical and dynamic settings. Dynamic Scene Graphs (DSG)~\cite{rosinol20203ddynamicscenegraphs} introduce multi-layer representations with spatio-temporal relations, while systems such as Kimera~\cite{rosinol2020kimera} and Hydra~\cite{hughes2022hydra} enable real-time construction of semantic graphs from sensor data. More recently, Khronos~\cite{schmid2024khronos} incorporates temporal evolution into metric-semantic SLAM. In structured environments, Situational Graphs (S-Graphs)~\cite{bavle2023sgraphs+} unify pose and semantic representations into a single optimizable graph.
Despite their expressiveness, these approaches mainly focus on representation and do not explicitly exploit semantic relations to constrain graph matching. In particular, object-to-structure relations are not leveraged to reduce ambiguity during correspondence estimation.

\subsection{Graph Matching}

Graph matching is typically formulated as selecting a set of geometrically consistent correspondences. Classical approaches rely on maximum clique formulations~\cite{bailey2000data} or robust estimation techniques~\cite{antonante2021outlier}. CLIPPER~\cite{lusk2021clipper} provides an efficient relaxation for correspondence selection and has shown strong performance in high-outlier regimes. Learning-based methods, such as NetVLAD~\cite{arandjelovic2016netvlad} and deep graph matching approaches~\cite{wang2019learning}, have also been explored.

\noindent\textbf{Scene graph matching.}
Matching scene graphs requires handling heterogeneous entities and their relations. MR-COGraphs~\cite{gu2025mr} matches graphs representing the objects observed in a scene by different robots, focusing on communication efficiency but lacking the use of the structure. Hydra~\cite{hughes2022hydra} performs hierarchical retrieval using multi-level descriptors followed by geometric verification, but semantic information is largely aggregated, limiting the use of explicit relational context. 
Graph-based localization approaches using S-Graphs~\cite{bavle2023sgraphs+} exploit topological and geometric consistency between rooms and planes, while diS-Graphs~\cite{shaheer2025tightly} extends this framework with incremental matching and deviation handling. However, these methods are restricted to structural elements (e.g., rooms and planes) and do not incorporate additional semantic cues from objects, which limits their ability to resolve ambiguities in symmetric environments.
Learning-based scene graph alignment has also been explored, e.g., SG-PGM~\cite{xie2024sg}, which fuses semantic and geometric features using graph neural networks. However, these approaches rely on learned representations rather than explicitly modeling semantic relations.

\section{System Overview}
\label{sec:system_overview}

% The proposed approach improves the robustness and efficiency of graph matching in structured indoor environments by introducing semantic relations between detected objects and structural elements. The key idea is to generate relations linking objects to elements such as rooms and wall planes, and to exploit these relations to reduce ambiguity during matching, particularly in environments with repeated layouts or symmetric structures.

% The semantic-enhanced matching framework is built on top of the \textit{diS-Graphs} pipeline \cite{shaheer2025tightly}, which extend S-Graphs by using architectural plans to provide accurate localization over the resulting hierarchical factor graphs. 
% extends Situational Graphs (S-Graphs) as a multi-layer, optimizable factor-graph representation that combines a pose graph with a scene graph enriched with semantic-relational concepts. Within this framework, an S-graph is built online during the inspection process from the robot observations, while an Architectural Graph (A-graph) is generated offline from Building Information Models (BIM) and used as a prior for place recognition and map alignment (Fig.~\ref{fig:main_diagram}).

The proposed system performs graph-based localization by matching an online graph (S-Graph) generated by the robot with a prior architectural graph (A-Graph). To improve the robustness and efficiency of this matching process, we incorporate semantic relations between detected objects and structural elements. 
The system comprises four main modules:
\begin{itemize}
    \item \textbf{S-graph generation:} A layered, hierarchical, optimizable factor graph is constructed and updated online during the inspection, representing the environment observed by the robot at multiple abstraction levels. This is generated using \textit{S-Graphs+} \cite{bavle2023sgraphs+}.
    \item \textbf{Enriched A-graph generation:} A layered, hierarchical prior graph is extracted from a BIM (Building Information Modeling) \cite{Azhar2008BuildingIM} file. The basic structural A-graph is enriched with new objects and relations that will be exploited during matching.
    \item \textbf{Object-semantic relation generator:} Each keyframe produced by the SLAM system is associated with an RGB-D frame used for visual object detection. Detected semantic-relevant objects are then linked to structural elements in the graphs (e.g., object inside a room, object on a wall plane), and the resulting relations are incorporated into the corresponding graph layers.
    \item \textbf{Semantic-enhanced graph matching:} The relations added to both the S-graph and the A-graph are exploited to filter candidate correspondences prior to geometric matching. This semantic filtering reduces the correspondence search space, improving the robustness when geometric structure alone is ambiguous, and reducing the computation time. Finally, geometric matching exploits the graph structure, node attributes, and geometric information of the remaining candidates to find a unique correspondence between the graphs. 
    \item \textbf{Graph Coupling:} Unique matches are forwarded to the \textit{iS-Graphs} module to integrate the S-graph with the A-graph for localization.
\end{itemize}

% \tb{I don't know where to put this}
% Although objects could be incorporated as additional nodes in the graph, the proposed design focuses on explicitly generating relations between objects and structural elements. This avoids increasing graph size and reduces the dependence on highly accurate object localization, while still providing informative cues for place recognition. Moreover, defining relations directly from object geometry would require a consistent room reference frame (e.g., a unique room orientation), which is generally unknown before correspondence is established.

The following sections describe the system components in detail: the object--semantic relation generator (Section~\ref{sec:objsem_relation_generator}) and the semantic-enhanced graph matching procedure (Section~\ref{sec:enhanced_graph_matching}).

\subsection{Object-semantic relation generator}
\label{sec:objsem_relation_generator}

\subsubsection{Object detection}
\label{sec:object_detector}

The object detection is performed on RGB-D data by translating visual detections into a semantically segmented 3D point cloud. The detector processes the keyframes provided by the \textit{S-Graphs} SLAM system using the RGB and depth images associated with each keyframe. Performing detection on keyframes reduces redundant processing while ensuring that objects are detected from distinct viewpoints already integrated in the SLAM representation. 
Visual object detection is applied to the RGB images, and the resulting detections are projected onto the associated depth point cloud. This projection produces a semantically segmented point cloud in which points within the detected regions are labeled according to the corresponding object category. The resulting detections are therefore directly linked to the corresponding keyframe.

\subsubsection{Data association}

Each detected object instance is represented using an ellipsoid that approximates its spatial extent. Ellipsoids provide a compact geometric abstraction that is robust to partial observations and enables efficient geometric reasoning, such as overlap or proximity checks. Since the objective is to establish semantic relations with structural elements rather than reconstruct precise object geometry, high accuracy in the object pose is not required, making this lightweight representation sufficient for our purposes.
Ellipsoids are estimated from the semantically labeled point cloud. Points belonging to each detected object category are processed independently by first extracting the corresponding subset of the labeled point cloud and then clustering it to separate individual object instances. For each cluster, an ellipsoid is estimated from the cluster points, approximating a single object instance observation. The point cloud used to generate each ellipsoid is also stored to allow incremental updates.
To perform data association across observations, newly generated ellipsoids are compared with previously detected ellipsoids of the same object category. If a new ellipsoid overlaps with an existing one, the corresponding point clouds are merged and a new ellipsoid is estimated from the combined points. Otherwise, the ellipsoid is stored as a new object instance.

% \begin{figure}[thpb]
%       \centering
%       \includegraphics[width=0.5\linewidth]{images/relation_creation.png}
%       \caption{Generation of room–object relations using visibility and plane constraints.}
%       \label{fig:roomobjectrelation}
% \end{figure}

\subsubsection{Doorways}
\label{sec:doorways}
Visual detection of doors and doorways can be challenging. To address this issue, we introduce an additional doorway detection strategy based on the robot trajectory. Using the keyframes provided by the SLAM system, the approach described in Section \ref{sec:object_in_rooms} is used to determine the room associated with each keyframe. Keyframes corresponding to the last robot pose inside a room and the first pose outside that room (or vice versa) are identified as doorway keyframes. The segment connecting these poses is then used to determine which plane of the room has been crossed by the robot. Finally, a doorway object is localized at the intersection between this segment and the corresponding plane, and the semantic relation is created.
This approach exploits the semantic role of doorways as structural openings that separate adjacent rooms and as free passages that allow the robot to move from one room to another.

% \tb{We generate an ellipsoid for this doorway. This also allows us to detect the doorways without visual information.}

\subsubsection{Relation generator}

The detected objects are used to generate semantic relations between the objects and the rooms and planes already present in the baseline system. 
The selection of these relations is motivated by the characteristics of indoor construction and inspection scenarios. In such environments, the number of detectable objects is typically limited compared to general indoor scenes. Therefore, we focus on structural or semi-structural elements that are expected to remain in fixed locations, such as doors and windows. These elements are also commonly represented in BIM, making them particularly suitable for establishing consistent semantic correspondences between the prior model and the observations.

% Two types of relations are defined: 

% \begin{itemize}
% \item \textbf{Objects in rooms:} a relation between a room node and an object node, used for general objects that can be located anywhere inside the room.
% \item \textbf{Objects on wall planes:} a relation between a specific plane of a room and an object node, used for structural elements such as doors and windows that are attached to room boundaries.
% \end{itemize}

Two types of relations are defined: objects in rooms and objects on wall planes.
These relations are incorporated into the original graph to produce an object-semantic-enriched graph.
By defining relations between rooms and the objects contained within them, as well as between room planes and the objects attached to them, the graph captures meaningful semantic cues that complement the geometric structure of the environment.

\textbf{Objects in rooms:}
\label{sec:object_in_rooms}
Rooms are defined by the set of planes that represent their boundary walls. Let $\mathcal{P} = \{\pi_i\}_{i=1}^{N}$ denote the set of planes of a room, where each plane is defined as
% \[
$\pi_i : \mathbf{n}_i^\top \mathbf{x} + d_i = 0$,
% \]
with $\mathbf{n}_i$ the plane normal and $d_i$ the plane offset. To determine whether an object belongs to a room, we perform a visibility test from the room center. Let $\mathbf{c}_r$ denote the room center and $\mathbf{c}_o$ the object centroid. A ray from the room center to the object is defined as
\[
\mathbf{r}(t) = \mathbf{c}_r + t(\mathbf{c}_o - \mathbf{c}_r), \quad t \ge 0.
\]
The intersection between the ray and plane $\pi_i$ occurs at
\[
t_i =
-\frac{\mathbf{n}_i^\top \mathbf{c}_r + d_i}
{\mathbf{n}_i^\top (\mathbf{c}_o - \mathbf{c}_r)}.
\]

If a plane intersects the ray before reaching the object, i.e., if $0 < t_i < 1$, the object is considered occluded. Otherwise, the object is visible from the room center and is therefore classified as belonging to the room. In practice, a small margin parameter $\epsilon$ is introduced and the condition $0 < t_i < 1-\epsilon$ is used to account for small errors in the estimated object position and prevent incorrect occlusions when the object lies close to a room plane. 
% Fig.~\ref{fig:roomobjectrelation} illustrates this procedure.

This criterion is adopted because rooms are represented by infinite planes that do not encode their spatial extent. Instead of estimating finite plane boundaries, we rely on a visibility-based approximation, which is valid for convex room geometries.

% This visibility-based test is adopted because rooms are represented only by their boundary planes, which are modeled as infinite planes. Since these planes do not encode their spatial extent, determining whether a point lies inside the room would require estimating the finite boundaries of each plane. Instead, we approximate room membership using a visibility test from the room center: if the ray connecting the room center and the object centroid is not intersected by any boundary plane before reaching the object, the object is considered to lie inside the room. This criterion is valid for convex room geometries, where any point inside the room remains visible from the interior.

\textbf{Objects on wall planes:}
Certain objects, such as doors and windows, are expected to lie on room planes. 
% \tb{A subset of objects is included in this category.} 
For this subset of objects, once they are classified as belonging to the room, we identify the closest plane using the point-to-plane distance
% \[
$d_i(\mathbf{c}_o) = \mathbf{n}_i^\top \mathbf{c}_o + d_i$.
% \]
The associated plane is selected as
\[
i^* = \arg\min_i |d_i(\mathbf{c}_o)|.
\]

If the distance satisfies $|d_{i^*}(\mathbf{c}_o)| < \tau$, where $\tau$ is a predefined threshold, the object is considered to lie on that plane, and a relation between the object and the plane node is added to the graph.
% If the distance satisfies
% \[
% |d_{i^*}(\mathbf{c}_o)| < \tau,
% \]
% where $\tau$ is a predefined threshold, the object is considered to lie on that plane and a relation between the object and the plane node is added to the graph.

%%%%%%%%%%%%%%%%%%%%%%%%%%%%%%%%%%%%%%%%%%%%%%%%%%%%%%%%%%%%%%%%%%%%%%

\subsection{Semantic-enhanced graph matching}
\label{sec:enhanced_graph_matching}

% To perform place recognition and map alignment, we match the graph generated online with a prior graph using a geometric graph matching approach. The baseline system represents the environment as a hierarchical graph composed of rooms and their boundary planes. In this work, we extend this representation by incorporating the semantic information described in the previous section.

% Specifically, the room–plane geometric graph is enriched with object nodes corresponding to the detected semantic elements. These additional nodes and edges produce an object-semantic-enriched graph that jointly encodes the geometric structure of the environment and the semantic context.

% The matching process operates on this enriched graph. 
Candidate correspondences between nodes of the A-graph and S-graph are first generated using categorical compatibility. Before performing geometric consistency evaluation, we introduce a semantic filtering stage that removes candidate matches inconsistent with the object content associated with rooms and planes, discarding geometrically plausible but semantically inconsistent matches. The remaining candidates are then evaluated using the hierarchy-based geometric matching procedure described in \cite{shaheer2023graph}, with the extensions detailed in the following section.

By incorporating these semantic relations before the geometric matching stage, the search space of possible correspondences is reduced, improving the robustness of the matching process, especially in environments with repeated geometric structures. This semantic filtering step significantly reduces the number of candidate correspondences that must be evaluated during the geometric consistency stage. 

% \tb{This new information also provides orientation cues that help disambiguate symmetric configurations. For example, the position of doors or windows on room boundary planes defines a directional reference that can be used to align corresponding rooms between the graphs. These semantic cues complement the geometric constraints and reduce ambiguities in environments with repeated or symmetric structures.}

\subsubsection{Hierarchy-based geometric matching}
The base algorithm~\cite{shaheer2023graph} performs a top-down candidate search between an A-Graph and an S-Graph by leveraging their hierarchical structure, accounting for node type, graph topology, and geometric consistency against a set of thresholds by stage.
Combinations are first constructed at the room level, and only those passing geometric consistency checks are extended to incorporate plane-level nodes.
Consistency is then evaluated between room-plane pairs, and subsequently across all possible plane combinations within those rooms.
Consistent candidates are aggregated across multiple rooms and their corresponding planes, covering both graphs in their entirety.
When the consistency difference between the two top-ranked candidates falls below a threshold, the presence of symmetries is assumed, and the decision of a unique match is deferred until additional observations populate the S-Graph.

% The base method also addresses deviations in plane positions between the S-Graph and the A-Graph by first tightening the geometric threshold until a unique match of deviation-free planes is found, then relaxing the criterion to incorporate deviated planes that remain consistent with that match.
\subsubsection{Object-semantic Filtering} 
We introduce a semantic filtering stage at two different levels of the previously explained base algorithm. When room-room or plane-plane potential candidates are searched, those that do not comply with the object-semantic condition are filtered out. 
Let $\mathcal{C}(n)$ denote the semantic content of node $n$, represented as a mapping from object categories to the corresponding detected instances. Given a candidate correspondence $(u,v)$ between an A-graph node $u$ and an S-graph node $v$, the match is accepted only if every object category detected in $v$ is also present in $u$, and the number of detected instances in $u$ is not smaller than in $v$, i.e.,
\[
\forall c \in \text{dom}(\mathcal{C}(v)), \quad c \in \text{dom}(\mathcal{C}(u)) \quad \wedge \quad \mathcal{C}(u)[c] \ge \mathcal{C}(v)[c].
\]
This condition reflects the fact that the online graph is typically incomplete, whereas the prior graph may contain a more complete semantic description. Therefore, we require semantic containment rather than exact equality. Candidate pairs that do not satisfy this condition are discarded before the geometric matching stage.

% \tb{JA suggestion: At the room and plane levels, each node is associated with the set of detected object instances contained in that region. Let $\mathcal{C}(n)$ denote the semantic neighborhood of node $n$, represented as a mapping from object categories to instance counts, where $\mathcal{C}(n)[c]$ denotes the number of detected instances of category $c$ associated with node $n$. Given a candidate correspondence $(u, v)$ between a prior node $u$ and an online node $v$, the match is accepted only if, for every object category present in $v$, the same category is present in $u$ with an equal or greater instance count, i.e.,}
% \[
% \forall c \in \mathcal{K}(v), \qquad c \in \mathcal{K}(u)
% \quad \wedge \quad
% |O_u^c| \ge |O_v^c|.
% \]

% \begin{figure}[thpb]
%       \centering
%       \includegraphics[width=0.9\linewidth]{images/sim_scenario_numbers.png}
%       \caption{Simulation environment used for the experiments. The scenarios were created using a subset of rooms with the order shown by the numbers.}
%       \label{fig:sim_scenario}
% \end{figure}

\section{Experimental Validation}

\subsection{Methodology}

We evaluate how the use of semantic relations improves the efficiency and robustness of the system. 
% For that the following experiments are going to be carried out:

\subsubsection{Matching performance} To assess the effect of the semantic filtering in the matching process, a synthetically generated dataset is used to analyze the intrinsic behavior of the matching algorithm independently of the perception pipeline. 
The following metrics are evaluated:

\begin{itemize}
\item \textbf{Computation time}: execution time of the graph matching algorithm when attempting to find all valid correspondences between the detected graph and the prior map.
\item \textbf{Number of solutions}: number of valid candidate correspondences. A unique match is obtained when only one solution remains, so higher values indicate greater ambiguity in the matching problem due to symmetries or lack of information.
% \item \textbf{Rooms until match}: number of detected nodes required to obtain a unique match.
% \item \textbf{Candidate reduction ratio}: ratio between the number of candidate correspondences before and after semantic filtering.
\end{itemize}

% \subsubsection{Convergence behavior} To understand the behavior of the system in a robotic application, the convergence behavior of the system during exploration was evaluated in simulation and real-world experiments using the following metrics:

% % % For simulation and real-world experiments, we focus on the convergence behavior of the system during exploration:
% \begin{itemize}
% \item \textbf{Convergence time}: time required to obtain the first unique match with the prior map.
% \item \textbf{Detection performance}: Avg precision, recall and F1 metrics for the objects in the environment (windows, doors, and both al them mixed).
% \item \textbf{ATE?}:
% \item \textbf{Rooms until convergence}: number of rooms detected before obtaining a valid match.
% \end{itemize}

\begin{figure}[thpb]
      \centering
      \includegraphics[width=1.0\linewidth]{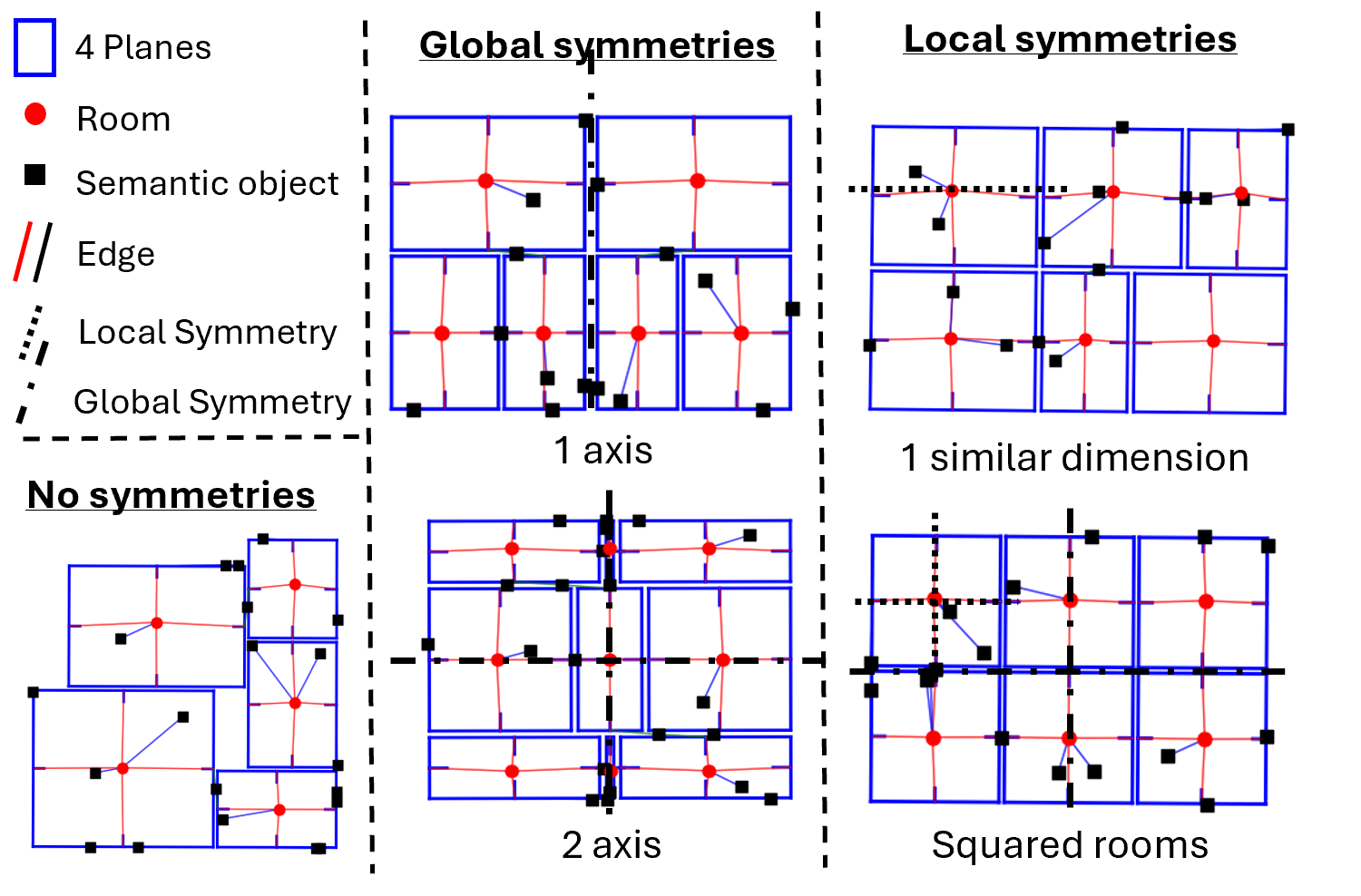}
      \caption{Samples of simulated layouts with rectangular rooms, in increasing matching complexity from unambiguous to global (full layout) and local (room-wise) symmetries.}
      \label{fig:synthetic_layouts}
      \vspace{-10pt}
\end{figure}

\subsubsection{Exploration performance}
To evaluate the behavior of the system in realistic robotic scenarios, we analyze its performance during exploration in a simulation environment. Since the proposed approach relies on semantic information extracted from detected objects, we also evaluate the perception performance to assess its impact on the overall system.
The following metrics are considered:

\begin{itemize}
\item \textbf{Convergence time}: time required to obtain the first unique match during exploration.
\item \textbf{Detection performance}: average precision, recall, and F1-score for object detection (windows, doors, and their combined evaluation).
\end{itemize}

% The performance of the proposed approach is evaluated by comparing it with the baseline geometric matching used in previous versions of the system \tb{and related state-of-the-art approaches.}

% Experiments are conducted in three settings: (i) synthetic data for controlled ablation studies, (ii) simulated environments to analyze the behavior of the system under symmetric layouts, and (iii) real-world datasets collected during inspection missions. Each experiment focuses on different aspects of the proposed method and therefore uses different evaluation metrics. Synthetic experiments focus on the intrinsic behavior of the matching algorithm, simulation experiments evaluate performance in symmetric layouts, and real-world experiments validate the approach in practical inspection scenarios.

\subsection{Datasets}
% Experiments are conducted in three settings: Synthetic experiments focus on the intrinsic behavior of the matching algorithm, simulation experiments evaluate performance in symmetric layouts, and real-world experiments validate the approach in practical inspection scenarios.
\subsubsection{Synthetic dataset}
Several random synthetic layouts with different numbers of rectangular rooms and objects. Settings like room size, wall thickness, plant size, position, and orientation are fully randomized to create realistic environments introduced in~\cite{reasoning_v1}. As shown in Fig.~\ref{fig:synthetic_layouts}, we randomly incorporate objects and apply symmetries at local (to all rooms) and at global level (to the whole layout).
In this setting, object detection and relation generation are not executed; instead, the semantic relations are directly generated within the synthetic graphs. This allows us to isolate the effect of semantic information on the graph matching process without introducing noise from perception or mapping components. 

% We used a synthetic generator data to generate random scenarios and get general measurements of the performance. The metrics are:
% * Computing matching time vs number of nodes detected and the number of nodes in the prior map.
% * Number of nodes needed to match.
% * Success rate
% * Candidate reduction ratio

\subsubsection{Simulation Data}

\begin{figure}[t]
      \centering
      \includegraphics[width=0.9\linewidth]{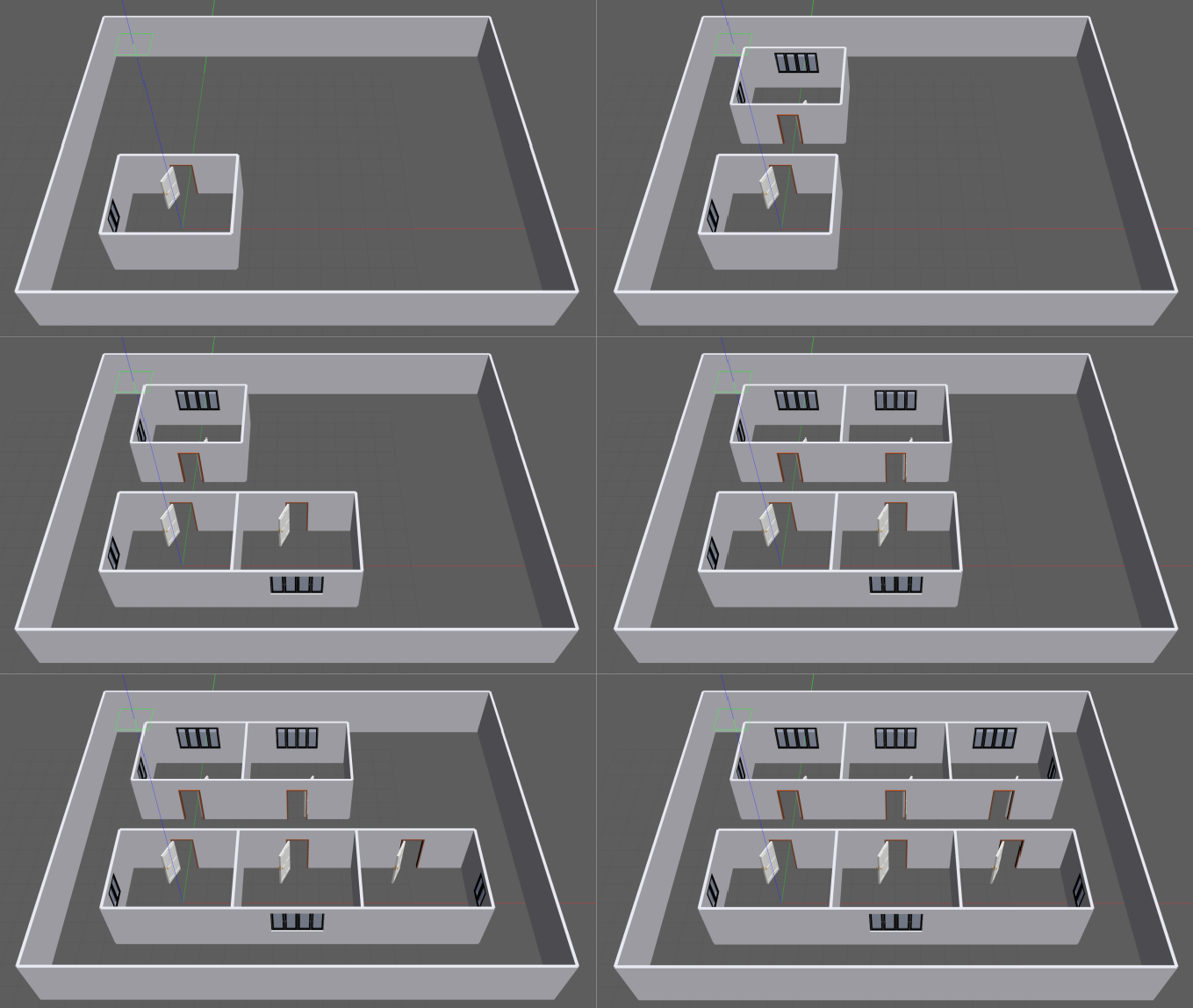}
      \caption{Simulated indoor environments. From top-left to bottom-right, the layouts evolve from simple configurations to multi-room scenarios, exhibiting different levels of symmetry. The rooms are inspected sequentially by the robot in the same order.}
      \label{fig:sim_scenarios}
      \vspace{-10pt}
\end{figure}

To evaluate the impact of semantic information on the convergence time of the graph matching, we designed a simulation experiment using a highly symmetric environment. Symmetric layouts are particularly challenging for purely geometric approaches, as several configurations may generate identical geometric structures and therefore lead to ambiguous matches.
A six-room layout was designed, and six scenarios were generated starting from only one room and incrementally adding one room at a time, as shown in Fig. \ref{fig:sim_scenarios}. This setup provides an interpretable scenario that connects the synthetic analysis with practical cases, where different map sizes and structural symmetries expose the limitations of purely geometric matching.
Since the spatial configuration of the objects in the simulation can be arbitrarily defined, measuring a single matching time does not provide a fair comparison between approaches. Instead, we analyze all possible matching opportunities as rooms are detected. For each scenario, we record the time ($T_i$) at which a different unique match can be established.
The experiment is performed using two configurations: (i) geometric-only matching and (ii) semantic-enhanced matching. This allows us to evaluate how semantic information helps resolve ambiguities caused by structural symmetry and enables earlier matching in the different scenarios.
The dataset was recorded in the Gazebo simulator \cite{koenig2004gazebo} using a simulated Boston Dynamics Spot robot equipped with a Velodyne VLP-16 LiDAR and an Intel RealSense D435 RGB-D camera.

% During the experiment, rooms are progressively detected over time, simulating the exploration of the environment. At each detection step ($T_i$), a new room becomes available and a matching attempt between the detected rooms and the map rooms is performed.

%%%%%%%%%% REAL WORLD EXPERIMENTS
% \subsubsection{Real-world Data}
% We used real data recordings collected at four construction sites (RE1 to RE4) with available architectural plans. These datasets were previously used in \cite{shaheer2025tightly}, where the baseline algorithms were compared with several state-of-the-art approaches \cite{}, enabling a direct comparison with previously reported results. Since the recordings do not contain RGB-D camera data, the only detectable objects are doorways, which are identified using the approach described in Section~\ref{sec:doorways}.

%%%%%%%%%%%%%%%%%%%%%%%%%%%%%%%%%%%%%%%%%%%%%%%%%%%%%%%%%%%%%%%

\subsection{Experimental setup}

% Real-world datasets were collected using the same robot–LiDAR configuration as in the simulation setup, but without the RGB-D camera.
% Simulated and real experiments were executed on a laptop equipped with an Intel® Core™ i7-14650HX processor (16 cores, 5.2 GHz) and 32 GB of RAM.
Experiments were conducted on a laptop equipped with an Intel® Core™ i9-12900H processor (14 cores, 20 threads, base frequency 2.9 GHz, turbo up to 5.0 GHz), and an NVIDIA T600 Laptop GPU with 4 GB VRAM (driver 535.183.01, CUDA 12.2).
% The proposed solution is implemented and evaluated within the \textit{diS-Graphs} framework \cite{shaheer2025tightly} using ROS~2.
We use YOSO \cite{hu2023yoso} as the object detector because it is trained to detect architectural elements such as doors, doorways, and windows, which are difficult to detect reliably with most general-purpose object detectors, where even open-world detection models often struggle to identify these elements consistently.

% For simulation, we used a dataset recorded in a Gazebo simulator \cite{koenig2004gazebo}, using a Boston Dynamics Spot simulated model with a Velodyne VLP-16 and a Realsense D435 device onboard.
% Real datasets were collected using the same robot-lidar configuration than in simulation but without an RGB-D camera.
% The experiments were carried out using a laptop computer with an Intel® Core™ i7-14650HX (16 cores, 5.2 GHz) with 32 GB of RAM memory.
% The presented solution is validated along with \textit{diS-graphs} \cite{shaheer2025tightly} using ROS 2.
% \tb{\textit{Multi S-Graphs} is developed under \textit{S-Graphs+}\cite{Bavle2022} using ROS 2 for implementing the algorithms. The room matching parameters have been tuned to give a robust performance in the experimental environments, obtaining values of $SC_{th}=0.35$ and $ICP_{th}=0.07$.}

%%%%%%%%%%%%%%%%%%%%%%%%%%%%%%%%%%%%%%%%%%%%%%%%%%%%%%%%%%%%%%%%%%

\begin{figure}[thpb]
      \centering
      \includegraphics[width=1.0\linewidth]{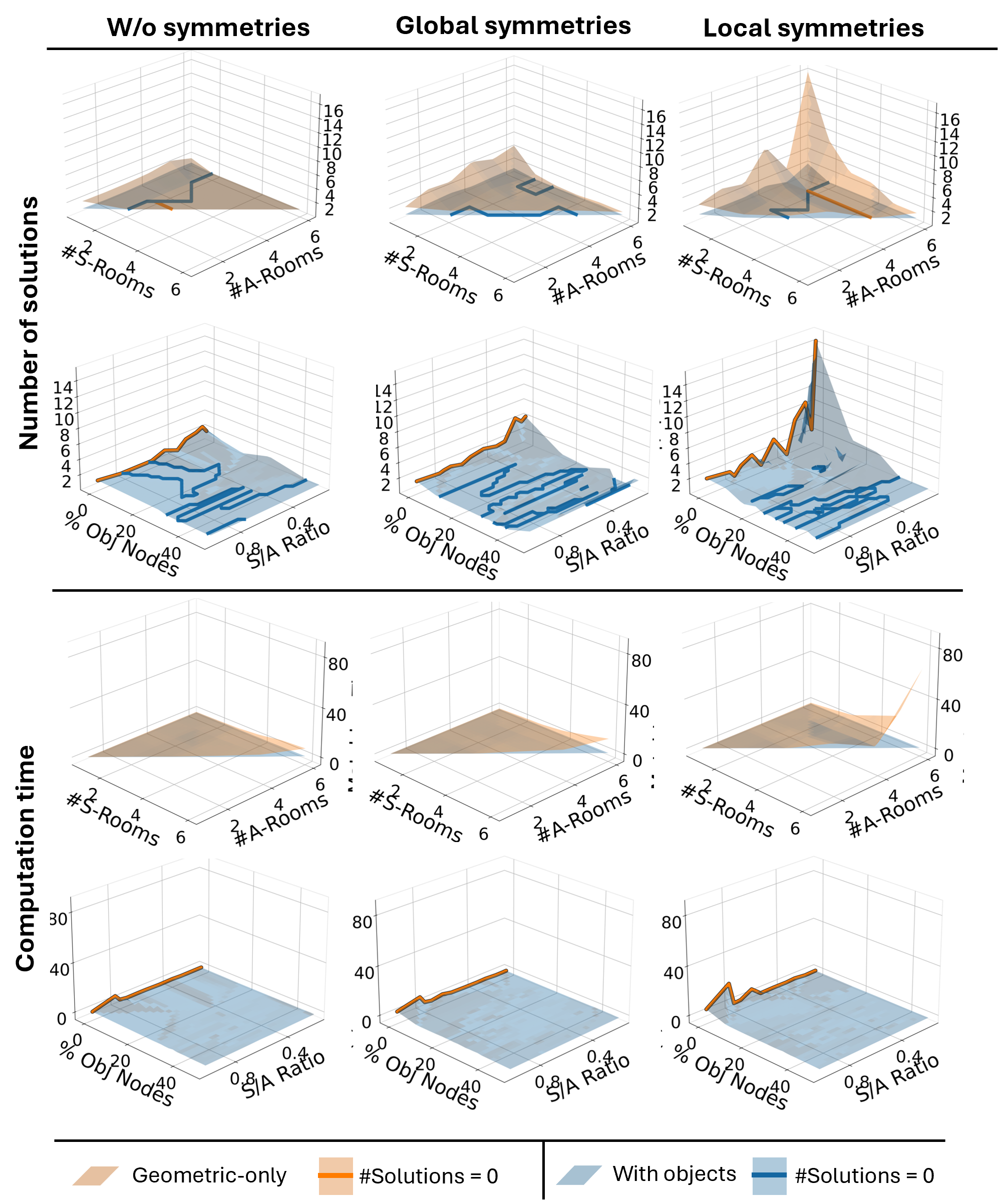}
      \caption{Comparison of the number of solutions found by each approach and the computation time across different symmetry types present in the layout. The surfaces are averaged on the vertical dimension and splined over the horizontal dimensions. \#A or S-Rooms is the number of rooms in the graph, S/A Ratio refers to the number of rooms. The percentage of object nodes is computed against the total number of nodes in the graph.}
      \label{fig:synthetic_metrics}
      \vspace{-10pt}
\end{figure}

\subsection{Results and Discussion}

\subsubsection{Matching performance}
% \textbf{Number of solutions}: 
Fig.\ref{fig:synthetic_metrics} (top two rows) reports the number of solutions found by both approaches across varying symmetry complexities. Values are averaged over layouts with the same room counts, so values close to 1 indicate that a unique match was recovered in most cases. Without object nodes (geometric-only, orange surface), a solution is almost never found. When objects are incorporated (blue surface), the surface settles at 1 solution across a broader region. The most challenging configurations, even when leveraging objects, are those with few rooms in the S-Graph: around 2 rooms suffice under no symmetry, around 3 are required under local symmetries, while global symmetries prevent a unique match over a wider region. The second row shows a clear tendency toward a unique match at approximately 20\% object density for higher room-count ratios, as expected. However, given the variability of the layouts and the disambiguation capacity of individual objects, stronger conclusions cannot be drawn.
% \textbf{Computation time}:
As shown in Fig.\ref{fig:synthetic_metrics} (bottom two rows), computation time generally increases with the number of rooms in both graphs, exhibiting an inflection point around 3–4 rooms. In layouts without symmetries, this growth remains approximately linear; global symmetries introduce a more pronounced increase, whereas local symmetries cause a sharper rise. Regarding the effect of object nodes (fourth row), their presence reduces computation time by pruning the candidate set, and a higher proportion of such objects does not incur additional overhead.

\subsubsection{Exploration performance}

Table \ref{tab:results_simulation_convergence} illustrates the matching availability as rooms are progressively detected. When semantic information is used, valid correspondences can be established before the full environment is observed. For example, in a three-room scenario the correct match can already be obtained just by visiting one or two rooms. In contrast, purely geometric matching fails in symmetric configurations and only succeeds when the room layout incidentally breaks the symmetry (e.g., at three and five rooms). Even in a scenario with a single room, the geometric information alone is insufficient to find a unique solution because it cannot determine the correct orientation of the room.
Table \ref{tab:detection_metrics} reports the object detection performance in 10 runs in the 6 rooms simulation scenario, showing precision and F1-scores above 0.95 across object categories. This indicates that the perception module provides reliable semantic information, making it well suited for the proposed approach.
Although the disposition of objects may also exhibit symmetric configurations, the additional information they provide significantly reduces the likelihood of such ambiguities. In our experiments, we deliberately used a limited set of object categories, focusing on structural elements commonly found in inspection scenarios. Nevertheless, the proposed system already allows the use of additional object categories, further increasing the discriminative power of the semantic graph and reducing the probability of symmetric matches.

\begin{table}[t]
\centering
\begin{tabular}{c|cccccc}
\hline
 & \multicolumn{6}{c}{Detected Rooms} \\
Map Rooms & 1 & 2 & 3 & 4 & 5 & 6 \\
\hline
\multicolumn{7}{c}{\textbf{Semantic Matching}} \\
\hline
1 & 41 &  &  &  &  &  \\
2 & 43 & 84 &  &  &  &  \\
3 & 41 & 87 & 147 &  &  &  \\
4 & 42 & 89 & 149 & 202 &  &  \\
5 & 43 & 90 & 152 & 210 & 270 &  \\
6 & 41 & 90 & 151 & 209 & 275 & 305 \\
\hline
\multicolumn{7}{c}{\textbf{Geometric Matching}} \\
\hline
1 & -- &  &  &  &  &  \\
2 & -- & -- &  &  &  &  \\
3 & -- & -- & 148 &  &  &  \\
4 & -- & -- & -- & -- &  &  \\
5 & -- & -- & -- & -- & 272 &  \\
6 & -- & -- & -- & -- & -- & -- \\
\hline
\end{tabular}
\caption{Convergence times (s) in simulation scenario. Each cell indicates that a valid match is available at time \(T_j\). Cells with '--' indicate that no valid match can be established.}
\label{tab:results_simulation_convergence}
\end{table}

\begin{table}[t]
\centering
\begin{tabular}{lccc}
\toprule
 & Window & Door & All Objects \\
\midrule
Avg Precision & 0.96 & 1.00 & 0.97 \\
Avg Recall    & 0.94 & 0.92 & 0.93 \\
Avg F1        & 0.95 & 0.96 & 0.95 \\
\bottomrule
\end{tabular}
\caption{Detection performance in the simulation scenario.}
\label{tab:detection_metrics}
\vspace{-10pt}
\end{table}

\section{Conclusions and Future Work}

This work presented a semantic-enhanced graph matching approach for indoor inspection scenarios. The proposed method extends the geometric scene graph representation by incorporating semantic relations between structural elements and detected objects. These relations are used to introduce a semantic filtering stage that reduces the number of candidate correspondences before geometric matching.

Experimental results demonstrate that this filtering significantly improves both efficiency and robustness. In synthetic experiments, the number of candidate solutions is drastically reduced, leading to faster matching without additional computational overhead. In simulation scenarios, the proposed approach enables successful matching before the full environment is observed, often requiring only partial exploration, while purely geometric methods fail in symmetric configurations.

% These results highlight that semantic relations provide critical disambiguation cues in environments with structural symmetries, where geometric and topological information alone is insufficient. Moreover, the use of a limited set of object categories already yields substantial improvements, suggesting that even sparse semantic information can significantly enhance graph matching.

Future work will focus on validation in real-world environments and incorporating a richer set of semantic entities.

\addtolength{\textheight}{-12cm}   % This command serves to balance the column lengths
                                  % on the last page of the document manually. It shortens
                                  % the textheight of the last page by a suitable amount.
                                  % This command does not take effect until the next page
                                  % so it should come on the page before the last. Make
                                  % sure that you do not shorten the textheight too much.

%%%%%%%%%%%%%%%%%%%%%%%%%%%%%%%%%%%%%%%%%%%%%%%%%%%%%%%%%%%%%%%%%%%%%%%%%%%%%%%%

%%%%%%%%%%%%%%%%%%%%%%%%%%%%%%%%%%%%%%%%%%%%%%%%%%%%%%%%%%%%%%%%%%%%%%%%%%%%%%%%

%%%%%%%%%%%%%%%%%%%%%%%%%%%%%%%%%%%%%%%%%%%%%%%%%%%%%%%%%%%%%%%%%%%%%%%%%%%%%%%%
% \section*{APPENDIX}

% Appendixes should appear before the acknowledgment.

% \section*{ACKNOWLEDGMENT}

% The preferred spelling of the word ÒacknowledgmentÓ in America is without an ÒeÓ after the ÒgÓ. Avoid the stilted expression, ÒOne of us (R. B. G.) thanks . . .Ó  Instead, try ÒR. B. G. thanksÓ. Put sponsor acknowledgments in the unnumbered footnote on the first page.

%%%%%%%%%%%%%%%%%%%%%%%%%%%%%%%%%%%%%%%%%%%%%%%%%%%%%%%%%%%%%%%%%%%%%%%%%%%%%%%%
% \input{bibliography}

\bibliographystyle{IEEEtran}
\bibliography{bibliography}
\end{document}